# COMPARATIVE ANALYSIS OF WORD EMBEDDINGS FOR CAPTURING WORD SIMILARITIES

Martina Toshevska[1], Frosina Stojanovska[2] and Jovan Kalajdjieski[3]

[1]Faculty of Computer Science and Engineering, Ss. Cyril and Methodius University, Skopje, Macedonia
`martina.toshevska@finki.ukim.mk`

[2]Faculty of Computer Science and Engineering, Ss. Cyril and Methodius University, Skopje, Macedonia
`frosina.stojanovska@finki.ukim.mk`

[3]Faculty of Computer Science and Engineering, Ss. Cyril and Methodius University, Skopje, Macedonia
`jovan.kalajdzhieski@finki.ukim.mk`

## ABSTRACT

*Distributed language representation has become the most widely used technique for language representation in various natural language processing tasks. Most of the natural language processing models that are based on deep learning techniques use already pre-trained distributed word representations, commonly called word embeddings. Determining the most qualitative word embeddings is of crucial importance for such models. However, selecting the appropriate word embeddings is a perplexing task since the projected embedding space is not intuitive to humans.*

*In this paper, we explore different approaches for creating distributed word representations. We perform an intrinsic evaluation of several state-of-the-art word embedding methods. Their performance on capturing word similarities is analysed with existing benchmark datasets for word pairs similarities. The research in this paper conducts a correlation analysis between ground truth word similarities and similarities obtained by different word embedding methods.*

## 1. INTRODUCTION

Representing natural language has become one of the main concerns in a wide range of Natural Language Processing (NLP) tasks. Some of the simplest techniques for representing natural language, when the data is textual, are one-hot encoding, Bag of Words (BoW), term-frequency, along with others [1]. These methods have a deficiency of incorporating the necessary information in the mapped space of the representations. Representations such as one-hot encoding create vector representations with enormous size (the size of the vocabulary) that do not preserve the relative closeness of the words, and representations such as BoW do not capture the context and order of the words.

Distributed representations of textual data as vectors have been proven useful in many algorithms. Each word vector in a given corpus is represented based on the mutual information with other words in the corpus [2]. Vector representation can be computed at different levels including: characters [3], [4], words [5], [6], [7], phrases [8], [9], sentences [10], document [11], etc. Word embeddings are representations of words in a continuous $R^n$ space. Each word is projected in n-dimensional vector space as a real-valued vector by keeping the information about the meanings and similarities of the words. Words with similar meanings will be mapped closer in the projected space. Word vectors capture syntactic and semantic regularities in language, such that the operation "King - Man + Woman" results in a vector close to "Queen" [5], [12].

Nowadays, several word embeddings methods incorporate different techniques for word representation. Training the word embeddings requires large corpus, extensive computational power and time. Consequently, the preferred approach is to use the idea of transfer learning, i.e. using already pre-trained embeddings and then training the NLP model with these embeddings. However, selecting the appropriate word embeddings is a perplexing task since the projected embedding space is not intuitive to humans. Nevertheless, deciding on the pre-trained word embeddings used as input in the NLP model is the first and crucial step, since the quality of the word representation has a significant influence on the global performance of the model.

Therefore, offering comparative experimental results for the methods can be valuable to researchers for determining the most appropriate embeddings for their model based on the comparative analysis. The evaluation of word embeddings is not unified. Generally, the evaluation methods can be mainly grouped into two different approaches: an intrinsic and extrinsic evaluation [13]. Intrinsic evaluations test the quality of a representation independent of specific natural language processing tasks, directly testing the syntactic or semantic relationships between words. Human-created benchmark datasets generated with a particular task of comparing words by similarity, relatedness, or similar, are of crucial importance in the comparing process of intrinsic evaluation. Extrinsic evaluation of word vectors is the evaluation of the embeddings on a real NLP task, chosen as an evaluation method, for example, part-of-speech tagging [14], named-entity recognition [15], sentiment analysis [16] and neural machine translation [17].

Determining the most qualitative word embeddings is the main focus when executing the evaluations. Additionally, the dimensionality of the word embeddings is an important segment of the method capability to encode notable information in the projected space. Bigger dimensionality means more space for incorporating more information in the space, but after reaching some point, the marginal gain will diminish [5]. Typically, the dimensionality of the vectors is set to be between 50 and 1,000, where 300 is the most frequently used. Consequently, evaluating the word embeddings requires additional investigation for finding the most suitable dimensionality.

In this paper, we focus on different pre-trained word embeddings used in state-of-the-art models for some NLP tasks and analyse their performance on capturing word similarities with existing benchmark datasets for word pairs similarities[1]. The datasets WordSim353 [18], SimLex999 [19] and SimVerb3500 [20] are used in the experiments to compare the state-of-the-art word embeddings.

The rest of the paper is organized as follows. Section 2 describes the word embedding methods used for comparative analysis. Section 3 gives an overview of the related research for the problem. The comparative analysis along with the discussion of the results of the experiments are included in Section 4. Finally, Section 5 concludes the paper and gives future work ideas.

---

[1] The code is available at https://github.com/mtoshevska/Comparing-Embeddings

# 2. WORD EMBEDDINGS

Mainly, word embeddings methods can be divided into three groups according to the technique of creating the vectors: neural network based, word matrix based and a combination of the previous two groups with an additional strategy of supplementary information fusion called ensemble methods.

## 2.1. Neural Network Based Methods

### 2.1.1. Word2Vec

Word2Vec includes two different models: Continuous Bag Of Words (CBOW) and Skip-gram [5], [6]. Both of these methods are neural networks with a hidden layer with N neurons, where N is the dimensionality of the created word embeddings. The first method CBOW is a neural network where the context of the words is the input of the network. The task is to predict the current word as the output of the network [5]. The second method Skip-gram is a neural network where the input is the one-hot encoding of the word, and the output is the predicted context of the word, i.e. the surrounding words [5]. To make these algorithms more efficient techniques such as Hierarchical Softmax and Skip-Gram Negative Sampling (SGNS) are used.

The most popular pre-trained Word2Vec word embeddings are 300-dimensional vectors generated with negative sampling training on the Google News corpus (that contains 100 billion words)[2] [6]. These pre-trained embeddings are used in the experiments.

### 2.1.2. FastText

The FastText model [21], [22] is directly derived from the Skip-gram model of Word2Vec. The authors claim that by using a distinct vector representation for each word, the Skip-gram model ignores the internal structure of the words. For that purpose, they suggested a different scoring function that takes into account the internal structure. Their subword model represents each word as a bag of character n-gram. Special symbols < and > are added at the beginning and the end of words to distinguish prefixes and suffixes from other character sequences. The word is also included in the set of its n-grams to be able to learn a better representation for each word. The authors suggest extracting all n-grams for n greater than or equal to 3 and smaller than or equal to 6.

This simple approach provides extraction of all prefixes and suffixes of a given word. When all the n-grams of a specific word are extracted, a vector representation is assigned to every n-gram. The final word is represented as the sum of the vector representations of all of its n-grams. This model allows sharing the representations across words, thus allowing to learn a reliable representation for rare words.

To bound the memory requirements, a hashing function - FNV-1a[3] is used to map the n-grams to integer numbers. There are also proposed approaches for producing compact architectures [23].

In our experiments, we have used pre-trained models both trained with subword information on Wikipedia 2017 (16B tokens) and trained with subword information on Common Crawl (600B tokens)[4].

---

[2] https://code.google.com/p/word2vec/, last visited: 02.04.2020

[3] http://www.isthe.com/chongo/tech/comp/fnv/, last visited: 02.04.2020

## 2.2. Word Matrix Based Methods

### 2.2.1. GloVe

GloVe (Global Vectors for Word Representation) [7] is a log-bilinear regression model for unsupervised learning of word representations that combines the advantages of two model families: global matrix factorization and local context window methods. The general idea is that co-occurrence ratio of any two words, that is the frequency of the words occurring in each other's context, encodes information about the words. It captures meaningful linear substructures, efficiently leveraging global word-word co-occurrence matrix statistics. The model is optimized so that the dot product of any pair of vectors is equal to the co-occurrence ratio of the corresponding words.

There are several pre-trained GloVe vectors available for public use[5], trained on different datasets such as Twitter and Wikipedia. In our experiments we include 50-dimensional and 200-dimensional word vectors pre-trained on the Twitter dataset (27B tokens), and 50-dimensional and 300-dimensional word vectors pre-trained on the Wikipedia dataset (6B tokens).

### 2.2.2. LexVec

LexVec [24] is based on the idea of factoring the PPMI matrix [25] using a reconstruction loss function. This loss function does not weigh all errors equally, unlike SVD, but penalizes errors of frequent co-occurrence more heavily while still treating negative co-occurrence, unlike GloVe. The authors propose keeping the SGNS [6] weighting scheme by using window sampling and negative sampling but explicitly factoring PPMI matrix rather than implicitly factorizing the shifted PMI matrix. The minimization on the two terms of the loss function is done using two approaches: mini-batch, which executes gradient descent in the same way as SGNS, and stochastic, where every context window is extended with k negative samples and iterative gradient descent is then run on pairs for each window. The authors state that the evaluation of word similarity and analogy tasks shows that LexVec compares to and often outperforms state-of-the-art methods on many of these tasks.

Several pre-trained LexVec vectors are available for public use[6], and the experiments in this paper include the model trained with subword information on Common Crawl which contained 2,000,000 words (58B tokens) and the model trained on Wikipedia which contained 368,999 words (7B tokens).

## 2.3. Ensemble Methods

### 2.3.1. ConceptNet Numberbatch

ConceptNet Numberbatch represents an ensemble method that produces word embeddings from large, multilingual vocabulary with a combination of the GloVe and Word2Vec embeddings and additional structured knowledge from the semantic networks ConceptNet [26] and PPDB [27]. This method uses the extended retrofitting technique [28] to adjust the pre-trained word embedding matrices with knowledge graphs. The objective is to learn a new embeddings matrix such that the word embeddings are close by some distance metric to their counterparts in the primary word embedding matrix and adjacent vertices in the knowledge network.

---

[4] https://fasttext.cc/docs/en/english-vectors.html, last visited: 02.04.2020
[5] https://nlp.stanford.edu/projects/glove/, last visited: 02.04.2020
[6] https://github.com/alexandres/lexvec, last visited: 02.04.2020

The pre-trained embeddings version ConceptNet Numberbatch 19.08[7] [29] is used in the experiments in this paper.

## 3. RELATED WORK

There are several attempts to compare different methods for creating word embeddings on different datasets for different purposes. Mainly the experiments are conducted on datasets containing English words, but some are on other languages as well. Berardi et al. [30] have conducted a comparison between the Skip-gram model of Word2Vec [5], [6] and GloVe [7] in the Italian language. The models are trained on two datasets: the entire dump of the Italian Wikipedia and a collection of 31,432 books (mostly novels) written in Italian. The authors believe that by using two very different datasets, both by purpose and style, they will investigate the impact of the training data on the two models. In this research, they conclude that Word2Vec's Skip-gram model outperforms GloVe on both datasets. However, it does not have the same accuracy as trained on English datasets, which may be a sign of a higher complexity of the Italian language.

Comparison between different models for word embeddings in different languages is also conducted in [31], where the authors first compare Hungarian analogical questions to English questions by training a Skip-gram model on the Hungarian Webcorpus. They obtain similar results in the morphological questions, but the semantic questions are dominated by the English model. They also conduct a proto dictionary generation and comparison using CBOW Word2Vec and GloVe models on Hungarian/Slovenian/Lithuanian languages to English.

In contrast, there have been quite a few studies evaluating word embeddings in quantitatively representing word semantics in the English language, mostly comparing the word embeddings generated by different methods. Baroni et al. [32] conducted a comparison between four models: Word2Vec's CBOW[8], DISSECT[9], Distributional Memory model[10] and Collobert and Weston[11]. The evaluation has been done using a corpus of 2.8 billion tokens. In their research, they concluded that Word2Vec's CBOW model outperformed other methods for almost all the tasks.

Ghannay et al. [33] evaluated the word embeddings generated by CSLM word embeddings [34], dependency-based word embeddings [35], combined word embeddings and Word2Vec's Skip-gram model on multiple NLP tasks. The models were trained on the Gigaword corpus composed of 4 billion words, and the authors found that the dependency-based word embeddings gave the best performance. They also concluded that significant improvement can be obtained by a combination of embeddings. Authors in [13] compared Word2Vec's CBOW model, GloVe, TSCCA [36], C&W embeddings [37], Hellinger PCA [38] and Sparse Random Projections [39] and concluded that Word2Vec's CBOW model outperformed the other encodings on 10 out of the 14 test datasets. Apart from comparisons of embeddings in the general NLP domain, there have also been comparisons of word embeddings on many specific domains such as biomedical domains [40], [41].

---

[7] https://github.com/commonsense/conceptnet-numberbatch, last visited: 02.04.2020
[8] https://code.google.com/p/word2vec/, last visited: 02.04.2020
[9] http://clic.cimec.unitn.it/composes/, last visited: 02.04.2020
[10] http://clic.cimec.unitn.it/dm/, last visited: 02.04.2020
[11] http://ronan.collobert.com/senna/, last visited: 02.04.2020

# 4. COMPARATIVE ANALYSIS

Word vectors can be evaluated using two different approaches. Extrinsic evaluation is an evaluation with real NLP tasks such as natural language inference or sentiment analysis. For that purpose, word vectors are incorporated into an embedding layer of a deep neural network. The model is then trained on a specific NLP task. If the performances are bad, it is uncertain whether the word vectors lack to represent the meaning of the words or simply the model is not good enough for the task.

The prior issue justifies the need for intrinsic evaluation. This type of evaluation is an intermediate evaluation where word vectors are directly evaluated on different tasks. The intrinsic evaluation involves tasks as word vector analogies and word similarities.

Word vector analogies are used to evaluate word vectors by how well their cosine distance after addition captures intuitive semantic and syntactic analogy questions. For example, the operation "King - Man + Woman" results in a vector close to "Queen", the operation "Windows - Microsoft + Google" results in a vector close to "Android", etc.

Word similarities are values that measure the similarity of word pairs. These values are collected in several human-generated datasets that are applied as a resource for comparison of the quality of mapping the words in some space with different word representations. Different datasets represent the similarity differently, by either coupling or segregating the word similarity with word relatedness or word association.

## 4.1. Data

WordSim353[12] dataset consists of 353 noun pairs with human similarity ratings [18]. The dataset is used as a golden standard in the field of word similarity and relatedness computation. The similarity scores of the word pairs in this resource are a representative measure for the similarity of the words dependent on the relatedness or association. Later the dataset is annotated with semantic relations by distinguishing the concepts of similarity and relatedness [42].

SimLex999[13] has recently become a widely used lexical resource for tracking progress in word similarity computation. This benchmark dataset is concentrated on measuring the similarity, rather than relatedness or association [19]. It consists of 666 noun-noun pairs, 222 verb-verb pairs and 111 adjective-adjective pairs. The annotation guidelines of the dataset narrow the similarity to synonymy [43]. SimLex999 can be used to measure the ability of the word embedding models to capture the similarity of the words (without the influence of the relatedness or association) while at the same time excluding the dependence of relatedness or association.

SimVerb3500[14] represents a dataset for verb similarity that consists of 3500 verb-verb pairs [20]. This dataset is built on the same annotation criteria as the SimLex-999, i.e. there is a difference between similarity and relatedness. Therefore, this method is a broad coverage resource for verb similarities.

---

[12] http://alfonseca.org/eng/research/wordsim353.html, last visited: 02.04.2020
[13] https://fh295.github.io/SimLex-999.zip, last visited: 02.04.2020
[14] https://www.repository.cam.ac.uk/handle/1810/264124, last visited: 02.04.2020

## 4.2. Word Similarity Experiments

### 4.2.1. Word Similarities

Word embeddings represent words in a way that words with similar meaning are represented with similar vectors. The experiments presented in this section evaluate the ability to capture word similarities of different pre-trained word embeddings. For each word pair in the datasets, the cosine similarity of the corresponding word embedding vectors is computed. Average cosine similarities for each dataset are plotted in Figure 1.

Average similarity obtained by human ratings is 5.86, 4.56 and 4.29 for WordSim353, SimLex999 and SimVerb3500 dataset, respectively. The closest average cosine similarity for each dataset is obtained among GloVe and FastText word embeddings. The average cosine similarities for GloVe embeddings is 5.37, 4.62 and 3.79 for WordSim353, SimLex999 and SimVerb3500 dataset, respectively. With the FastText embeddings, average cosine similarity is 4.69, 4.81 and 4.12 for WordSim353, SimLex999 and SimVerb3500 dataset, respectively. These values direct to the conclusion that FastText and GloVe perform better in capturing similarities between words.

In terms of dimensionality, it is expected that word vectors of higher dimensionality have a more prominent quality of the representation. This statement is confirmed for SimLex999 dataset and slightly confirmed for SimVerb3500 dataset, when analysing the GloVe embeddings with different dimensionalities. For SimLex999 dataset, average cosine similarity is closer to ground truth similarity for 200-dimensional embeddings pre-trained on Twitter and 300-dimensional embeddings pre-trained on Wikipedia. For SimVerb3500 dataset, the statement applies only for 200-dimensional embeddings pre-trained on Twitter, while the similarities for vectors pre-trained on Wikipedia are closer to ground truth similarities when the dimensionality is 50. For the WordSim353 dataset, lower vector dimensionality implies closer cosine to ground-truth similarity.

### 4.2.2. Correlation Analysis

Average similarity does not always provide good insight into word similarities since similarities obtained by one metric can be higher than similarities obtained by another metric. Despite different average values, both similarity distributions could be correlated. Therefore, for all pre-trained word embedding vectors, we calculate the correlation coefficient between cosine similarities of word vectors and ground truth similarities.

Spearman correlation coefficient, Pearson correlation coefficient and Kendall's tau correlation coefficient are computed for each pair of ground-truth similarities and cosine similarities. The results are summarized in Table 1, Table 2 and Table 3 for WordSim353, SimLex999 and SimVerb3500 dataset, respectively, and are depicted in Figure 2 for the WordSim353 dataset, Figure 3 for the SimLex999 dataset, and Figure 4 for the SimVerb3500 dataset.

Results of the correlation analysis oppose the findings of average similarity. GloVe and FastText word embeddings demonstrated better performance for capturing word similarities in terms of average cosine similarity. However, they have the lowest correlation coefficient. The value is around 0, implying no correlation between ground truth and cosine similarities of word vectors. FastText despite being developed to overcome weaknesses of Word2Vec, does not show to capture word similarities better since correlation is roughly 0.1 for SimLex999 and SimVerb3500 and approximately 0.2 for WordSim353. GloVe also shows weak performance on SimLex999 and SimVerb3500, although it is used as a starting point in many high-performance deep learning models for different NLP tasks. However, the correlation on WordSim353 is quite higher, with a maximum value of 0.61 for the Spearman correlation coefficient.

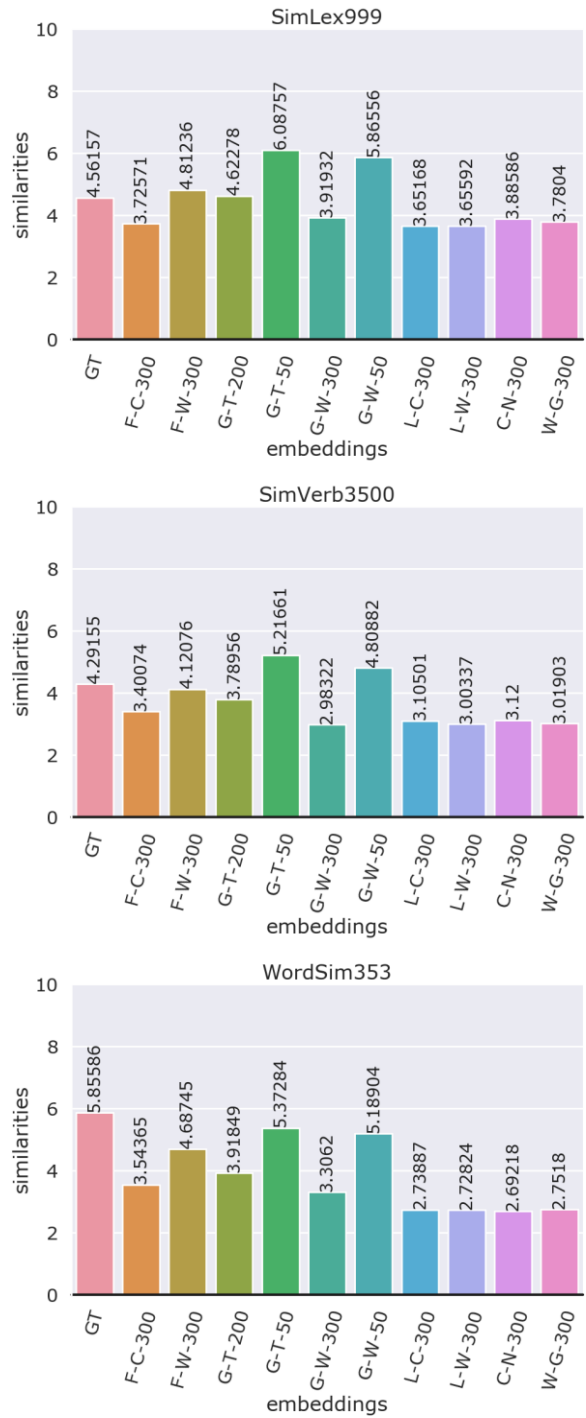

Figure 1. Average cosine similarities. GT - ground truth similarities. F-C-300 - 300-dimensional FastText embeddings pre-trained on Crawl dataset. F-W-300 - 300-dimensional FastText embeddings pre-trained on Wikipedia dataset. G-T-200 - 200-dimensional GloVe embeddings pre-trained on Twitter dataset. G-T-50 - 50-dimensional GloVe embeddings pre-trained on Twitter dataset. G-W-300 - 300-dimensional GloVe embeddings pre-trained on Wikipedia dataset. G-W-50 - 50-dimensional GloVe embeddings pre-trained on Wikipedia dataset. L-C-300 - 300-dimensional LexVec embeddings pre-trained on Crawl dataset. L-W-300 - 300-dimensional LexVec embeddings pre-trained on Wikipedia dataset. N-C-300 - 300-dimensional ConceptNet Numberbatch embeddings. W-G-300 - 300-dimensional Word2Vec embeddings pre-trained on Google News dataset.

For all three datasets, the highest correlation is achieved by ConceptNet Numberbatch word embeddings, indicating a positive correlation between similarity distributions. The second highest correlation coefficient belongs to Word2Vec and LexVec word embeddings that are about 0.7 for WordSim353 dataset, around 0.4 for SimLex999 dataset and roughly 0.3 for the SimVerb3500 dataset.

The assumption about the better quality of word vectors with higher dimensionality corresponds with the correlation results. Correlation coefficients for GloVe word embeddings despite being low for the SimLex999 and SimVerb3500 dataset, show that higher dimensionality entails higher correlation. Furthermore, we can infer that word vectors of higher dimensionality are better for capturing word similarities.

Table 1. Correlation coefficients between ground truth similarities and word vector cosine similarities for the WordSim353 dataset. S - Spearman correlation coefficient. Pr - Pearson correlation coefficient. K - Kendall's tau correlation coefficient.

| Word Embeddings | S | Pr | K |
| --- | --- | --- | --- |
| FastText-Crawl-300 | 0.25 | 0.26 | 0.17 |
| FastText-Wikipedia-300 | 0.19 | 0.19 | 0.13 |
| GloVe-Twitter-200 | 0.52 | 0.53 | 0.36 |
| GloVe-Twitter-50 | 0.46 | 0.46 | 0.32 |
| GloVe-Wikipedia-300 | 0.61 | 0.60 | 0.45 |
| GloVe-Wikipedia-50 | 0.50 | 0.51 | 0.36 |
| LexVec-Crawl-300 | 0.72 | 0.68 | 0.53 |
| LexVec-Wikipedia-300 | 0.66 | 0.63 | 0.48 |
| ConceptNet-Numberbatch-300 | 0.81 | 0.75 | 0.63 |
| Word2Vec-GoogleNews-300 | 0.69 | 0.65 | 0.51 |

Table 2. Correlation coefficients between ground truth similarities and word vector cosine similarities for the SimLex999 dataset. S - Spearman correlation coefficient. Pr - Pearson correlation coefficient. K - Kendall's tau correlation coefficient.

| Word Embeddings | S | Pr | K |
|---|---|---|---|
| FastText-Crawl-300 | 0.16 | 0.16 | 0.11 |
| FastText-Wikipedia-300 | 0.09 | 0.07 | 0.06 |
| GloVe-Twitter-200 | 0.13 | 0.14 | 0.08 |
| GloVe-Twitter-50 | 0.10 | 0.10 | 0.06 |
| GloVe-Wikipedia-300 | 0.37 | 0.39 | 0.30 |
| GloVe-Wikipedia-50 | 0.26 | 0.29 | 0.18 |
| LexVec-Crawl-300 | 0.44 | 0.45 | 0.31 |
| LexVec-Wikipedia-300 | 0.38 | 0.39 | 0.27 |
| ConceptNet-Numberbatch-300 | 0.63 | 0.65 | 0.46 |
| Word2Vec-GoogleNews-300 | 0.44 | 0.45 | 0.31 |

Table 3. Correlation coefficients between ground truth similarities and word vector cosine similarities for the SimVerb3500 dataset. S - Spearman correlation coefficient. Pr - Pearson correlation coefficient. K - Kendall's tau correlation coefficient.

| Word Embeddings | S | Pr | K |
|---|---|---|---|
| FastText-Crawl-300 | 0.11 | 0.11 | 0.07 |
| FastText-Wikipedia-300 | 0.03 | 0.02 | 0.02 |
| GloVe-Twitter-200 | 0.06 | 0.07 | 0.04 |
| GloVe-Twitter-50 | 0.03 | 0.04 | 0.02 |
| GloVe-Wikipedia-300 | 0.23 | 0.23 | 0.16 |
| GloVe-Wikipedia-50 | 0.15 | 0.16 | 0.10 |
| LexVec-Crawl-300 | 0.30 | 0.31 | 0.21 |
| LexVec-Wikipedia-300 | 0.28 | 0.28 | 0.19 |
| ConceptNet-Numberbatch-300 | 0.57 | 0.59 | 0.41 |
| Word2Vec-GoogleNews-300 | 0.36 | 0.38 | 0.25 |

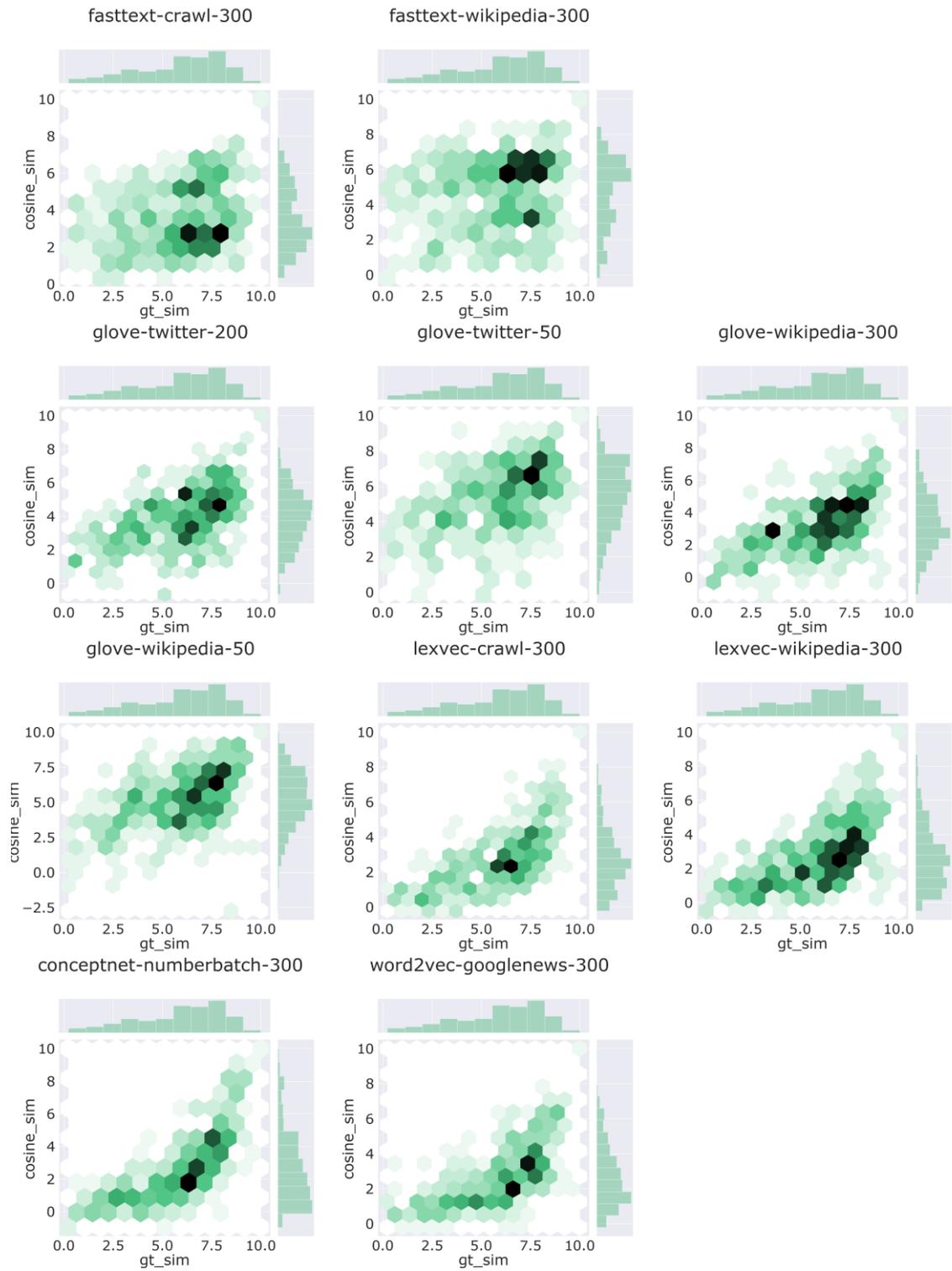

Figure 2. Correlation between ground truth similarities and similarities obtained with different word embedding methods for the WordSim353 dataset. Gt_sim – ground truth similarities obtained by human ratings. Cosine_sim – cosine similarities between word embeddings

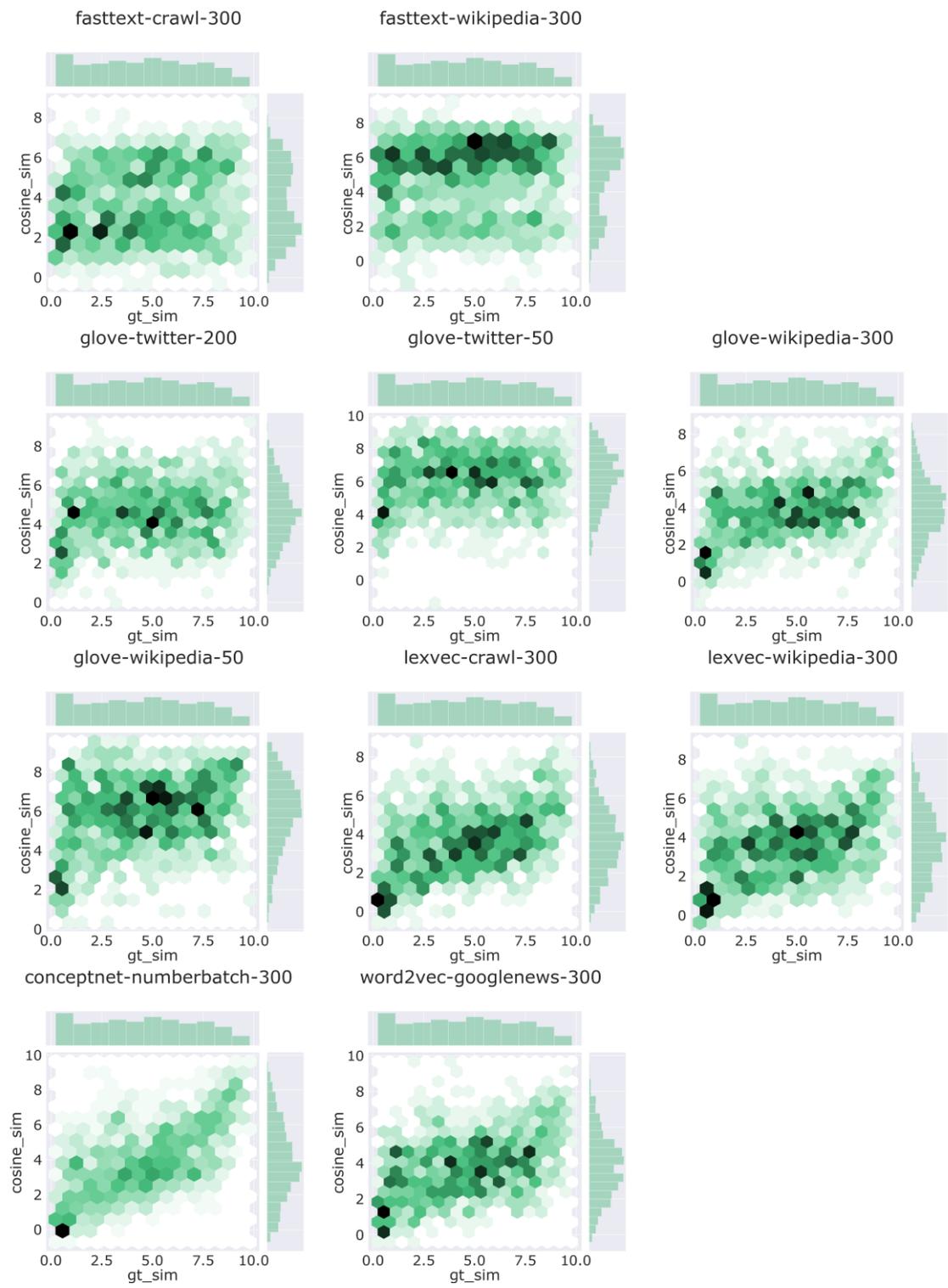

Figure 3. Correlation between ground truth similarities and similarities obtained with different word embedding methods for the SimLex999 dataset. Gt_sim – ground truth similarities obtained by human ratings. Cosine_sim – cosine similarities between word embeddings.

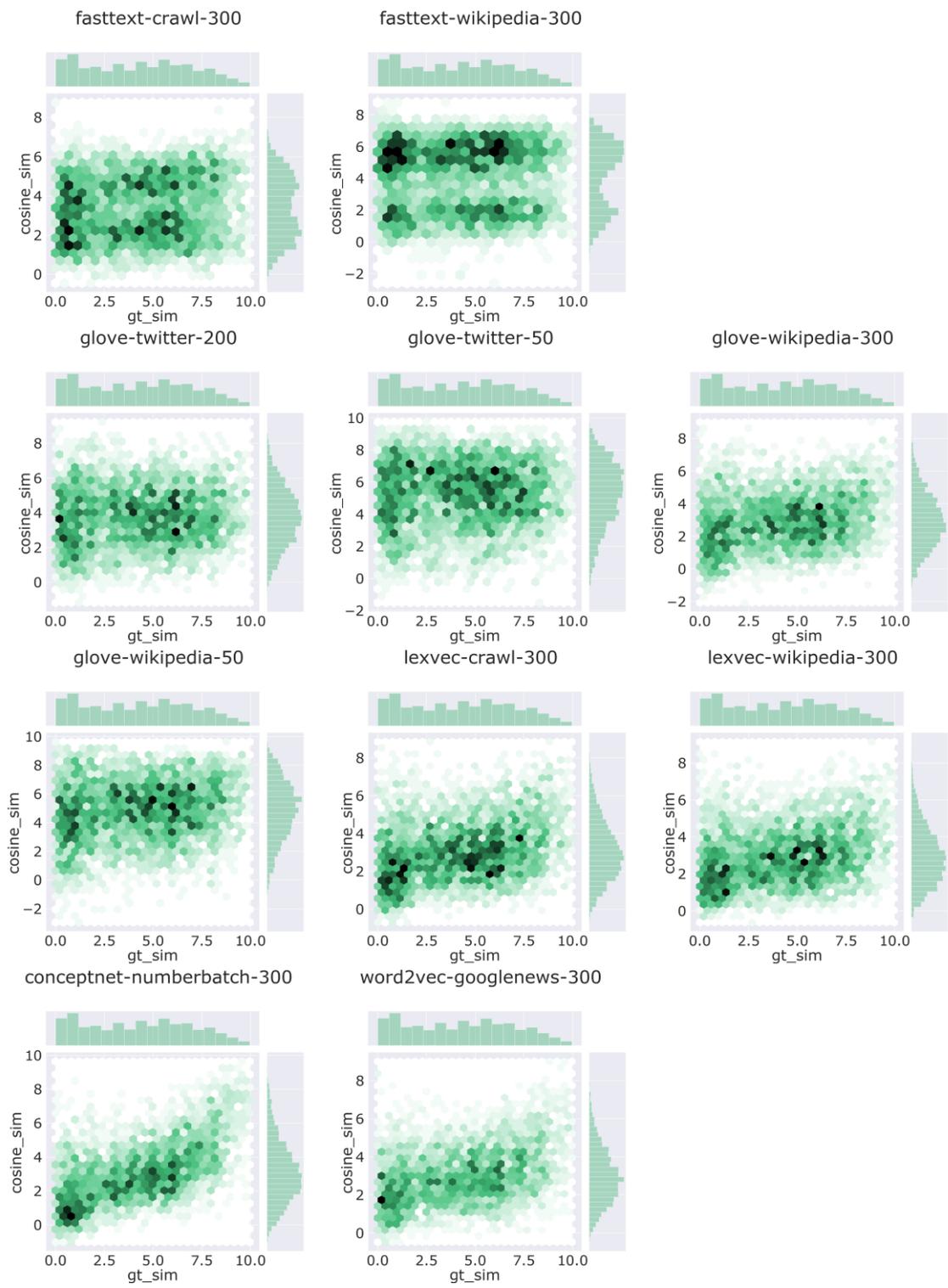

Figure 4. Correlation between ground truth similarities and similarities obtained with different word embedding methods for the SimVerb3500 dataset. Gt_sim – ground truth similarities obtained by human ratings. Cosine_sim – cosine similarities between word embeddings

## 5. CONCLUSIONS AND FUTURE WORK

In this paper, we have analysed several word embedding methods. These methods could be generally divided into two groups according to the creation technique: neural network based and matrix based. We further examined an ensemble method which is a compound of the two methods.

Word embedding similarity experiments were conducted for all of the examined methods. These experiments evaluate the ability to capture word similarities between pairs of words on three different datasets using a cosine similarity measure. Analysing only the average similarity information and comparing these values with the average value of the human ratings, the conclusion was that the GloVe and FastText outperformed the other word embedding methods.

After conducting the word embedding similarity experiments, we also carried out correlation analysis. In this analysis, the Spearman correlation coefficient, Pearson correlation coefficient and Kendall's tau correlation coefficient were computed for each pair of ground truth similarities and cosine similarities of the word embeddings. Using the data obtained in this analysis we have concluded that even with the high cosine similarity values of GloVe and FastText, their correlation values are close to 0, implying no correlation between ground truth and cosine similarities of word vectors. Moreover, ConceptNet Numberbatch word embeddings have the highest correlation coefficients even though they did not have high average similarity values.

A more comprehensive study of word similarities is including the context of the text of the words. Newest word embedding methods, like ELMO [44] or BERT [45], utilize the context of the words as they appear in the text, creating deep contextualized word representations. Examining the difference between these different mappings of the words will be part of our future work. Additionally, combining the different ideas from intrinsic and extrinsic evaluations would provide a more beneficial comparison of the different word representations.